\documentclass[10pt,twocolumn,letterpaper]{article}

\usepackage{iccv}
\usepackage{times}
\usepackage{epsfig}
\usepackage{graphicx}

\usepackage{amssymb,amsthm,amsmath,amscd}

\usepackage{epsfig}
\usepackage{subfigure}
\usepackage{algorithmic}
\usepackage{algorithm}
\usepackage{verbatim}

\iccvfinalcopy 

\def\iccvPaperID{43} 

\ificcvfinal\fi

\DeclareMathOperator{\dist}{dist}

\DeclareMathOperator{\Span}{span} 

\newcommand{\reals}{\mathbb R}

\newcommand{\be}{\begin{equation}}
\newcommand{\ee}{\end{equation}}

\newcommand{\di}{{\,\mathrm{d}}}
\def\cX{\mathrm{X}}
\def\bx{\mathbf{x}}
\def\by{\mathbf{y}}
\def\bz{\mathbf{z}}
\def\bP{\mathbf{P}}
\def\sX{\mathrm{X}}
\def \di{\mathrm{d}}

\begin{document}
%

\title{Median $K$-Flats for Hybrid Linear Modeling with Many Outliers \thanks{This work was supported by NSF grants \#0612608, \#0811203 and \#0915064}
}
\author{\begin{tabular}[t]{@{\extracolsep{\fill}}ccc}
Teng Zhang & Arthur Szlam & Gilad Lerman\\
School of Mathematics & Department of Mathematics   & School of Mathematics\\
University of Minnesota & University of California, LA & University of Minnesota\\
127 Vincent Hall &  Box 951555 & 127 Vincent Hall \\
206 Church Street SE &  Los Angeles, CA 90095 & 206 Church Street SE \\
Minneapolis, MN 55455 & {\small \tt aszlam@math.ucla.edu} & Minneapolis, MN 55455 \\
{\small \tt zhang620@umn.edu} &  & {\small \tt lerman@umn.edu}
\end{tabular}}
\date{today}
\maketitle
\begin{abstract}
We describe the Median $K$-flats (MKF) algorithm, a simple online
method for hybrid linear modeling, i.e., for approximating data by a
mixture of flats. This algorithm simultaneously partitions the data
into clusters while finding their corresponding best approximating
$\ell_1$ $d$-flats, so that the cumulative $\ell_1$ error is
minimized. The current implementation restricts $d$-flats to be
$d$-dimensional linear subspaces. It requires a negligible amount of
storage, and its complexity, when modeling data consisting of $N$
points in $\reals^D$ with $K$ $d$-dimensional linear subspaces, is
of order $O(n_s \cdot K \cdot d \cdot D+n_s \cdot d^2 \cdot D)$,
where $n_s$ is the number of iterations required for convergence
(empirically on the order of $10^4$). Since it is an online
algorithm, data can be supplied to it incrementally and it can
incrementally produce the corresponding output. The performance of
the algorithm is carefully evaluated using synthetic and real data.
\vspace{.1in}
\end{abstract}

\noindent \textbf{Supp. webpage}:
http://www.math.umn.edu/$\sim$lerman/mkf/

\section{Introduction}

Many common data sets can be modeled by mixtures of flats (i.e.,
affine subspaces). For example, feature vectors of different moving
objects in a video sequence lie on different affine subspaces (see
e.g., \cite{Ma07}), and similarly, images of different faces under
different illuminating conditions are on different linear subspaces
with each such subspace corresponding to a distinct
face~\cite{Basri03}. Such data give rise to the problem of hybrid
linear modeling, i.e., modeling data by a mixture of flats.

Different kinds of algorithms have been suggested for this problem
utilizing different mathematical theories. For example,
Generalized Principal Component Analysis (GPCA)~\cite{Vidal05} is
based on algebraic geometry, Agglomerative Lossy Compression
 (ALC)~\cite{Ma07Compression} uses information theory, and Spectral
Curvature Clustering (SCC)~\cite{spectral_applied} uses multi-way
clustering methods as well as multiscale geometric analysis. On the
other hand, there are also some heuristic
approaches, e.g., Subspace
Separation~\cite{Costeira98,Kanatani01,Kanatani02} and Local
Subspace Affinity (LSA)~\cite{Yan06LSA}. Probably, the most
straightforward method of all is the $K$-flats (KF) algorithm or any
of its
variants~\cite{Kambhatla94fastnon-linear,Tipping99mixtures,Bradley00kplanes,Tseng00nearest,Ho03}.

The $K$-flats algorithm aims to partition a given data set
$\cX = \{\bx_1, \dots, \bx_N\} \subseteq \reals^D$
into $K$ subsets $\cX_1, \ldots, \cX_K$,
each of which is well approximated by its best fit $d$-flat.
More formally, given parameters $K$ and $d$, the algorithm tries
to minimize the objective
function
\be \label{eq:objective_kflats}
 \sum_{i=1}^K \min_{d-\text{flats } L_i} \sum_{\bx_j \in \cX_i}
\dist^2(\bx_j,L_i)\,. \ee
In practice, the minimization of this function is performed
iteratively as in the $K$-means algorithm~\cite{MacQueen67}. That
is, after an initialization of $K$ $d$-flats (for example, they may be chosen
randomly), one repeats the following two steps until
convergence: 1) Assign clusters according to minimal distances to
the flats determined at the previous stage. 2) Compute least squares
$d$-flats for these newly obtained clusters by Principal
Component Analysis (PCA).

This procedure is very fast and is guaranteed to converge to at
least a local minimum.  However, in practice, the local minimum it
converges to is often significantly worse than the global minimum.
As a result, the $K$-flats algorithm is not as accurate as more
recent hybrid linear modeling algorithms, and even in the case of
underlying linear subspaces (as opposed to general affine subspaces)
it often fails when either $d$ is sufficiently large (e.g., $d \geq
10$) or there is a large component of outliers.

This paper has two goals. The first one is to show that in order to
significantly improve the robustness to outliers and noise of the
$K$-flats algorithm, it is sufficient to replace its objective
function (Eq.~\eqref{eq:objective_kflats}) with
\be \label{eq:objective_med_kflats} \sum_{i=1}^K \min_{d-\text{flats
} L_i} \sum_{\bx_j \in \cX_i} \dist(\bx_j,L_i)\,, \ee
that is, replacing the $\ell_2$ average with an $\ell_1$ average.
The second goal is to establish an online algorithm for this
purpose, so that data can be supplied to it incrementally, one point
at a time, and it can incrementally produce the corresponding
output. We believe that an online procedure, which has to be very
different than $K$-flats, can also be beneficial for standard
settings of moderate-size data which is not streaming. Indeed, it is
possible that such a strategy will converge more often to the global
minimum of the $\ell_1$ error than the straightforward $\ell_1$
generalization of $K$-flats (assuming an accurate algorithm for
computing best $\ell_1$ flats).

In order to address those goals we propose the Median $K$-flats
(MKF) algorithm. We chose this name since in the special case where
$d=0$ the well-known $K$-medians algorithm (see e.g., \cite{Jain88})
approximates the minimum of the same energy function.
The MKF algorithm employs a stochastic gradient descent
strategy~\cite{Bishop:Pattern} in order to provide an online
approximation for the best $\ell_1$ $d$-flats.
Its current implementation only applies to the setting of underlying
linear subspaces (and not general affine ones).

Numerical experiments with synthetic and real data indicate superior
performance of the MKF algorithm in various instances. In
particular, it outperforms some standard algorithms in the cases of
large outlier component or relatively large intrinsic dimension of
flats. Even on the Hopkins 155 Database for motion
segmentation~\cite{Tron2007}, which requires small intrinsic
dimensions, has little noise, and few outliers, the MKF performs
very well and in particular better than $K$-flats. We speculate that
this is because the iterative process of MKF converges more often to
a global minimum than that of the $K$-flats.

The rest of this paper is organized as follows. In Section~\ref{sec:MKF} we
introduce the MKF algorithm. Section~\ref{sec:exp} carefully tests the
algorithm on both artificial data of synthetic hybrid linear models
and real data of motion segmentation in video sequences.
Section~\ref{sec:conclusions} concludes with a brief discussion and mentions possibilities
for future work.

\section{The MKF algorithm}\label{sec:MKF}
We introduce here the MKF algorithm and estimate its storage and
running time. We then discuss some technical details of our implementation.

\subsection{Description of algorithm}

The MKF algorithm partitions a data set
$\mathrm{X}=\{\bx_1,\bx_2,\cdots,\bx_N\} \subseteq \mathbb{R}^{D}$
into $K$ clusters $\sX_1$, $\sX_2$, $\ldots$, $\sX_K$, with each
cluster approximated by a $d$-dimensional linear subspace.


We start with a notational convention for linear subspaces. For each
$1\leq i \leq K$, let $\bP_i$ be the $d \times D$ matrix whose rows
are the orthogonal basis of the linear subspace approximating
$\sX_i$, and note that $\bP_i\bP_i^T=\mathbf{I}_{d \times d}$. We identify
the approximating subspaces of clusters $\sX_1, \ldots, \sX_K$ with
the matrices $\bP_1, \ldots, \bP_K$.

We define the following energy function for the partition
$\{\sX_i\}_{i=1}^{K}$ and the corresponding subspaces
$\{\bP_i\}_{i=1}^{K}$:
\begin{equation}\label{eq:l1energy}
\mathcal{E}(\{\sX_i\}_{i=1}^{K},\{\bP_i\}_{i=1}^{K}) =\sum_{i=1}^K
\sum_{\bx\in \sX_i} ||\bx-\bP_i^T\bP_i \bx||.
\end{equation}
The MKF algorithm tries to partition the data into clusters
$\{\sX_i\}_{i=1}^{K}$ minimizing the above energy. Since the
underlying flats are linear subspaces, we can normalize the elements
of X to lie on the unit sphere, so that $||\bx_j||=1$ for each
$1\leq j \leq N$, and express the energy function $\mathcal{E}$ as
follows:
\begin{align}
\mathcal{E}(\{\sX_i\}_{i=1}^{K},\{\bP_i\}_{i=1}^{K}) &=\sum_{i=1}^K
\sum_{\bx\in \sX_i} \sqrt{||\bx-\bP_i^T \bP_i \bx||^2} \nonumber
\\
&=\sum_{i=1}^K \sum_{\bx\in \sX_i} \sqrt{1-||\bP_i \bx||^2}.
\end{align}

To minimize this energy, the MKF algorithm uses the method of
stochastic gradient descent~\cite{Bishop:Pattern}. The derivative of
the energy with respect to a given matrix $\bP_i$ is
\begin{equation}\frac{\partial \mathcal{E}
}{\partial \bP_i}=-\sum_{\bx\in \sX_i}
\frac{\bP_i\bx\bx^T}{\sqrt{1-||\bP_i \bx||^2}}.\end{equation}

The algorithm needs to adjust $\bP_i$ according to the component of
the derivative orthogonal to $\bP_i$.
The part of the derivative that is parallel to the subspace $\bP_i$ is
\begin{equation}
\frac{\partial
\mathcal{E}
}{\partial
\bP_i}\bP_i^T\bP_i=-\sum_{\bx\in \sX_i}
\frac{\bP_i\bx\bx^T\bP_i^T\bP_i}{\sqrt{1-||\bP_i \bx||^2}}.
\end{equation}
Hence the orthogonal component is
\begin{equation} \di\bP_i= \sum_{\bx\in \sX_i} \di_\bx \bP_i,
\end{equation}
where
\begin{equation}\label{eq:dxpi}
\di_\bx
\bP_i=-\frac{(\bP_i\bx\bx^T-\bP_i\bx\bx^T\bP_i^T\bP_i)}{\sqrt{1-||\bP_i
\bx||^2}}\,.
\end{equation}

In view of the above calculations, the algorithm proceeds by
picking a point $\bx^*$ at random from the set, and then deciding which
$\bP_{i^*}$ that point currently belongs to.  Then it applies the update
$\bP_{i^*}\mapsto \bP_{i^*} -\di t \di_{\bx^*}\bP_{i^*}$, where
$\di t$  (the ``time step'')
is a parameter chosen by the user.
It repeats this process until some
convergence criterion is met, and assigns the data points to their nearest subspaces
$\{\bP_i\}_{i=1}^K$ to obtain the $K$ clusters.  This is summarized in Algorithm~\ref{alg:theoretical}.
\begin{algorithm}
\caption{Median $K$-flats (MKF) \label{alg:theoretical} }
\begin{algorithmic}
\vspace{.1cm} \REQUIRE $\sX=\{\bx_1,\bx_2,\cdots,\bx_N\} \subseteq
\mathbb{R}^{D}$: data, normalized onto the unit sphere, $d$:
dimension of subspaces, $K$: number of subspaces,
$\{\bP_i\}_{i=1}^{K}$: the initialized subspaces. $\di t$: step
parameter.
\vspace{.1cm}
\ENSURE A partition of $\sX$ into $K$ disjoint clusters $\{\sX \}_{i=1}^K$.\\
\vspace{.1cm} \textbf{Steps}:
\STATE
\begin{enumerate}

    \item Pick a random point $\bx^*$ in $\sX$
    \item Find its closest subspace $\bP_{i^*}$, where
    $$i^*=\text{argmax}_{1 \leq i \leq K} ||\bP_{i}\bx||$$
    \item Compute $\di_{\bx^*} \bP_{i^*}$ by Eq.~\eqref{eq:dxpi}
    \item Update $\bP_{i^*}$: $\bP_{i^*}\mapsto \bP_{i^*} -\di t \di_{\bx^*}\bP_{i^*}$
    \item Orthogonalize $\bP_{i^*}$

\item
Repeat steps 1-5
until
convergence\setcounter{footnote}{\value{footnote}}\protect\footnotemark
\item  Assign each $\bx_i$ to the nearest subspace

\end{enumerate}
\end{algorithmic}
\end{algorithm}

\subsection{Complexity and storage of the algorithm}
Note that the data set does not need to be kept in memory, so the
storage requirement of the algorithm is $O(K \cdot d \cdot D)$, due
to the $K$ $d\times D$ matrices $\{\bP_i\}_{i=1}^{K}$.

Finding the nearest subspace to a given point costs  $O(K \cdot d
\cdot D)$ operations. Computing the update costs $O(d \cdot D)$, and
orthogonalizing $\bP_{i^*}$ costs $O(d^2 \cdot D)$. Consequently, each iteration
is $O(K \cdot d \cdot D+d^2 \cdot D)$.
If $n_s$ denotes the number of sampling iterations performed, then
the total running time of the MKF algorithm is $O(n_s \cdot K \cdot
d \cdot D+n_s \cdot d^2 \cdot D)$.

 In our experiments we use $\di t=0.01$.  With this choice, the number of sampling iterations
 $n_s$ is typically about $10^4$.
 Usually $n_s$ increases as the data becomes more complex (i.e., more flats, more outliers, etc),
 but in our experiments
 it never exceeded $3 \cdot 10^4$.

\footnotetext{In our experiments we checked the energy functional of
Eq.~\eqref{eq:l1energy} every 1000 iterations. We stopped if the
ratio between current energy and the previous one was in the range
(0.999,1.001). However, the computation of the energy functional
depends on the size of the data. For large data sets we can obtain
an online algorithm by replacing the ratio of the energy
functionals, with e.g., the sum of squares of sines of principal
angles between the corresponding subspaces.}

\subsection{Initialization}
Although the algorithm often works well with a random initialization
of $\{\bP_i\}_{i=1}^{K}$, it can many times be improved with a more
careful initialization. We propose a farthest insertion method in
Algorithm~\ref{alg:init} below.
\begin{algorithm}
\caption{Initialization for $\{\bP_i\}_{i=1}^{K}$ \label{alg:init} }
\begin{algorithmic}
\vspace{.1cm} \REQUIRE $\sX=\{\bx_1,\bx_2,\cdots,\bx_N)\} \in
\mathbb{R}^{D\times n}$: data, $d$: dimension, $K$: number of
$d$-flats
\vspace{.1cm}
\ENSURE $\{\bP_i\}_{i=1}^{K}$: $K$ subspaces.

\STATE
\vspace{.1cm} \textbf{For} $i=1$ to $K$, \textbf{do}
\begin{itemize}

\item If $i=1$, Pick a random point $\hat{\bx}$ in $\sX$; otherwise pick the point $\hat{\bx}$ with the largest distance from the available planes $\{P_1,P_2,\cdots,P_{i-1}\}$

\item Find the smallest integer $j$ such that
$$\dim(\Span(j \, \text{NN} (\hat{\bx}) -\hat{\bx}))=d,$$ where $j \,
\text{NN}(\hat{\bx})$ denotes the set of $j$-nearest neighbors of
$\hat{\bx}$

\item   Let $\bP_i$ be the affine space spanned by $\hat{\bx}$ and $j \,
\text{NN}(\hat{\bx})$

\end{itemize}

\textbf{end}

\end{algorithmic}
\end{algorithm}

If the data has little noise and few outliers, then empirically,
this initialization greatly increases the likelihood of obtaining
the correct subspaces.  On the other hand, in the case of
sufficiently large noise or outliers, the initialization of
Algorithm~\ref{alg:init} does not work significantly better than
random initializations, since the local structure of the data is
obscured.

Notice that the initialization of Algorithm~\ref{alg:init} also
works for affine subspaces, so we can use it to initialize other
iterative methods, such as $K$-flats.

\subsection{Some implementation odds and ends}

Because the algorithm is randomized and the objective function may
have many local minima, it is useful to restart the algorithm
several times as often practiced in the $K$-flats algorithm. We can
choose the best set of flats over all the restarts either measured
in the $\ell_1$ sense or in the $\ell_2$ sense, depending on the
application.

The MKF algorithm we have presented is designed for data sampled
from linear subspaces of the same dimension. 
For affine subspaces, similar as in~\cite{Vidal05} we can add a homogeneous coordinate so that subspaces become linear.
Empirically, it works well for clean cases with little noise or
few outliers. However, we are still working on the true affine model, to make the algorithm more accurate and robust.

Also, for mixed dimensions of subspaces, i.e., when the dimensions
$d_1$, $d_2$, $\cdots$, $d_K$ are not identical, we can set $d$ to
be $\max(d_1,d_2,\cdots,d_K)$ to implement the MKF algorithm
(similarly as in~\cite{spectral_applied}). Experiments show that
this method works well if there exists a comparably small difference
among $\{d_i\}_{i=1}^{K}$.

\section{Simulation and experimental results}
\label{sec:exp}
In this section, we conduct experiments on artificial and real data
sets to verify the effectiveness of the proposed MKF algorithm in
comparison to other hybrid linear modeling algorithms.

We measure the accuracy of those algorithms by the rate of
misclassified points with outliers excluded, that is
\begin{equation}
\text{error}\%= \frac{\text{\# of misclassified inliers}}{\text{\#
of total inliers}} \times 100\%\,.
\end{equation}

\subsection{Simulations}

\begin{table*}[htbp]
\centering \caption{{Mean percentage of misclassified points
in simulation. 
The MKF or KF algorithm with random initialization are denoted by
MKF(R) and KF(R) respectively. }\label{tab:error}}
\begin{tabular}{|l||r|r||r|r||r|r||r|r||r|r||r | r||r | r|}
 \hline
    &\multicolumn{2}{c||}{}
    &\multicolumn{2}{c||}{}
    &\multicolumn{2}{c||}{}
    &\multicolumn{2}{c||}{}
    &\multicolumn{2}{c||}{}
    &\multicolumn{2}{c||}{$(1,2,3)$}
    &\multicolumn{2}{c|}{$(4,5,6)$}\\
    \raisebox{1.5ex}[0pt]{\normalsize{Setting}}
    &\multicolumn{2} {c||}{\raisebox{1.5ex}[0pt]{$2^4 \in\mathbb{R}^4$}}
    &\multicolumn{2}{c||}{\raisebox{1.5ex}[0pt]{$4^2 \in\mathbb{R}^6$}}
    &\multicolumn{2}{c||}{\raisebox{1.5ex}[0pt]{$4^3 \in\mathbb{R}^6$}}
    &\multicolumn{2}{c||}{\raisebox{1.5ex}[0pt]{$10^2 \in\mathbb{R}^{15}$}}
    &\multicolumn{2}{c||}{\raisebox{1.5ex}[0pt]{$15^2 \in\mathbb{R}^{20}$}}
    &\multicolumn{2}{c||}{$\in\mathbb{R}^5$}
    &\multicolumn{2}{c|}{$\in\mathbb{R}^{10}$} \\

    \cline{1-15}
      Outl.  \%       & 5 & 30 & 5 & 30 & 5 & 30 & \;\;\; 5 & 30 & \;\;\;5 & 30 & 5 & 30 & 5 & 30 \\
  \hline\hline
GPCA&28.2&43.5&10.5&34.9&14.9&47.8&5.4&42.3&13.0&45.1&19.8&32.1&5.8&43.0\\
KF&7.8&30.2&2.2&15.4&4.8&27.7&0.6&34.8&2.2&43.4&9.1&25.2&0.8&26.7\\
KF(R)&8.3&32.8&2.2&15.9&4.8&30.8&0.5&28.8&2.2&41.7&11.0&26.3&0.9&25.4\\
LSA&42.6&46.1&10.6&12.0&21.1&26.5&7.0&8.9&13.1&16.6&29.6&31.3&5.8&6.7\\
LSCC&6.7&13.4&2.0&2.4&4.1&5.7&0.3&0.3&1.1&9.5&9.8&14.9&1.4&21.8\\
\textbf{MKF}&9.6&18.8&2.0&2.1&4.0&7.0&0.1&0.1&0.2&0.3&19.2&17.2&0.9&0.7\\
MKF(R)&7.6&17.6&2.0&2.0&3.9&9.7&0.2&0.1&0.2&0.3&17.6&17.1&1.1&0.7\\
MoPPCA&21.7&45.3&7.5&24.3&17.4&40.3&4.6&36.4&11.9&41.7&18.1&30.1&9.4&36.1\\

  \hline
\end{tabular}

\end{table*}
\begin{table*}[htbp]
\centering \caption{{Mean running time (in seconds) in
simulation.}\label{tab:time} }

\begin{tabular}{|l||r|r||r|r||r|r||r|r||r|r||r|r||r|r|}
 \hline
    &\multicolumn{2}{c||}{}
    &\multicolumn{2}{c||}{}
    &\multicolumn{2}{c||}{}
    &\multicolumn{2}{c||}{}
    &\multicolumn{2}{c||}{}
    &\multicolumn{2}{c||}{$(1,2,3)$}
    &\multicolumn{2}{c|}{$(4,5,6)$}\\
    \raisebox{1.5ex}[0pt]{\normalsize{Setting}}
    &\multicolumn{2} {c||}{\raisebox{1.5ex}[0pt]{$2^4 \in\mathbb{R}^4$}}
    &\multicolumn{2}{c||}{\raisebox{1.5ex}[0pt]{$4^2 \in\mathbb{R}^6$}}
    &\multicolumn{2}{c||}{\raisebox{1.5ex}[0pt]{$4^3 \in\mathbb{R}^6$}}
    &\multicolumn{2}{c||}{\raisebox{1.5ex}[0pt]{$10^2 \in\mathbb{R}^{15}$}}
    &\multicolumn{2}{c||}{\raisebox{1.5ex}[0pt]{$15^2 \in\mathbb{R}^{20}$}}
    &\multicolumn{2}{c||}{$\in\mathbb{R}^5$}
    &\multicolumn{2}{c|}{$\in\mathbb{R}^{10}$} \\
    \cline{2-15}

    \cline{1-15}
      Outli.  \%       & 5 & 30 & 5 & 30 & 5 & 30 &\;\;\; 5 & 30 & \;\;\;5 & 30 & 5 & 30 & 5 & 30 \\
  \hline\hline
GPCA&28.9&40.1&11.1&22.0&28.0&51.7&24.8&46.0&29.3&53.6&20.1&40.2&43.7&81.0\\
KF&1.3&1.6&0.3&0.6&0.8&1.4&0.7&1.0&1.3&1.3&0.7&1.2&1.0&1.8\\
KF(R)&1.7&1.8&0.5&0.8&1.1&1.7&0.7&1.2&1.4&1.6&0.8&1.4&1.0&1.9\\
LSA&47.7&92.1&13.0&24.8&30.6&59.8&25.5&47.9&31.6&59.5&28.7&56.7&43.2&82.3\\
LSCC&7.1&6.1&4.2&5.0&8.1&10.7&16.3&19.6&33.5&39.4&6.5&7.5&14.0&17.3\\
\textbf{MKF}&7.2&6.5&5.9&5.4&8.3&8.3&6.6&10.0&12.1&18.7&6.7&6.8&7.0&10.7\\
MKF(R)&7.7&6.7&6.5&5.9&9.2&8.6&9.0&12.4&15.3&20.7&7.5&7.3&10.5&13.6\\
MoPPCA&1.5&2.0&0.3&0.7&0.8&1.7&0.7&1.2&1.6&1.7&0.8&1.5&1.0&2.0\\

\hline
\end{tabular}
\vspace{.1in}
\end{table*}
\begin{table*}[htbp]
\centering \caption{{Standard deviation of misclassification rate in
simulation.}\label{tab:var} }\vspace{.1in}

\begin{tabular}{|l||r|r||r|r||r|r||r|r||r|r||r|r||r|r|}
 \hline

    &\multicolumn{2}{c||}{}
    &\multicolumn{2}{c||}{}
    &\multicolumn{2}{c||}{}
    &\multicolumn{2}{c||}{}
    &\multicolumn{2}{c||}{}
    &\multicolumn{2}{c||}{$(1,2,3)$}
    &\multicolumn{2}{c|}{$(4,5,6)$}\\
    \raisebox{1.5ex}[0pt]{\normalsize{Setting}}
    &\multicolumn{2} {c||}{\raisebox{1.5ex}[0pt]{$2^4 \in\mathbb{R}^4$}}
    &\multicolumn{2}{c||}{\raisebox{1.5ex}[0pt]{$4^2 \in\mathbb{R}^6$}}
    &\multicolumn{2}{c||}{\raisebox{1.5ex}[0pt]{$4^3 \in\mathbb{R}^6$}}
    &\multicolumn{2}{c||}{\raisebox{1.5ex}[0pt]{$10^2 \in\mathbb{R}^{15}$}}
    &\multicolumn{2}{c||}{\raisebox{1.5ex}[0pt]{$15^2 \in\mathbb{R}^{20}$}}
    &\multicolumn{2}{c||}{$\in\mathbb{R}^5$}
    &\multicolumn{2}{c|}{$\in\mathbb{R}^{10}$} \\
    \cline{1-15}

    \cline{1-15}
      Outli.  \%       & 5 & 30 & 5 & 30 & 5 & 30 &\;\;\; 5 & 30 & \;\;\;5 & 30 &5 & 30 &5 & 30 \\
  \hline\hline
GPCA&58.2&35.6&29.8&26.0&38.7&30.6&18.9&12.4&29.5&7.4&36.7&34.4&28.4&32.1\\
KF&19.6&35.9&2.5&26.4&10.8&27.9&0.9&21.0&2.5&10.9&21.0&45.2&1.4&37.0\\
KF(R)&21.1&34.4&2.5&25.9&10.7&28.7&0.9&22.2&2.6&13.5&27.4&41.4&3.6&40.5\\
LSA&21.7&23.8&8.1&8.1&15.3&19.2&3.9&4.5&4.5&4.9&21.7&20.6&5.1&5.7\\
LSCC&9.2&38.5&2.2&5.4&4.0&13.0&0.5&0.6&1.1&28.1&22.4&20.7&5.8&27.5\\
\textbf{MKF}&24.7&33.6&2.2&2.6&3.8&18.6&0.4&0.3&0.6&0.5&21.4&32.4&1.4&1.0\\
MKF(R)&16.9&36.5&2.1&2.5&3.8&29.9&0.4&0.3&0.5&0.5&21.5&30.2&2.0&1.1\\
MoPPCA&56.0&44.1&31.8&34.1&50.7&34.4&26.0&19.6&24.4&12.7&33.8&36.4&37.1&25.1\\

\hline
\end{tabular}
\vspace{.1in}
\end{table*}

We compare MKF with the following algorithms: Mixtures of PPCA
(MoPPCA)~\cite{Tipping99mixtures}, $K$-flats (KF)~\cite{Ho03}
(implemented for linear subspaces), Local Subspace Analysis
(LSA)~\cite{Yan06LSA}, Spectral Curvature Clustering
(SCC)~\cite{spectral_applied} (we use its version for linear
subspaces, LSCC) and GPCA with voting (GPCA)~\cite{Yang_voting,
Ma07}. We use the Matlab codes of the GPCA, MoPPCA and KF algorithms
from http://perception.csl.uiuc.edu/gpca, the LSCC algorithm from
http://www.math.umn.edu/$\sim$lerman /scc and the LSA algorithm from
http://www.vision.jhu.edu/db. The code for the MKF algorithm appears
in the supplementary webpage of this paper. It has been applied with
the default value of $\di t=0.01$.

The MoPPCA algorithm is always initialized with a random guess of
the membership of the data points. The LSCC algorithm is initialized
by randomly picking $100\times K$ $(d+1)$-tuples (following
~\cite{spectral_applied}). On the other hand, KF and MKF are
initialized with both random guess (they are denoted in this case by
KF(R) and MKF(R) respectively) as well as the initialization
suggested by Algorithm~\ref{alg:init} (and then denoted by KF and
MKF). We have used 10 restarts for MoPPCA, 30 restarts for KF, 5
restarts for MKF and 3 restarts for LSCC, and recorded the
misclassification rate of the one with the smallest $\ell_2$ error
(Eq.~\eqref{eq:objective_kflats}) for MoPPCA, LSCC as well as KF,
and $\ell_1$ error (Eq.~\eqref{eq:l1energy}) for MKF. The number of
restarts was restricted by the running time.


The simulated data represents various instances of $K$ linear
subspaces in $\mathbb{R}^D$. If their dimensions are fixed and equal
$d$, we follow~\cite{spectral_applied} and refer to the setting as
$d^K\in \mathbb{R}^D$. If they are mixed, then we follow~\cite{Ma07}
and refer to the setting as $(d_1, \ldots, d_K) \in \mathbb{R}^D$.
Fixing $K$ and $d$ (or $d_1, \ldots, d_K$), we randomly generate 100
different instances of corresponding hybrid linear models according
to the code in http://perception.csl.uiuc.edu/gpca. More precisely,
for each of the 100 experiments, $K$ linear subspaces of the
corresponding dimensions in $\reals^D$ are randomly generated.
Within each subspace the underlying sampling distribution is a cross
product of a uniform distribution along a $d$-dimensional cube of
sidelength 2 in that subspace centered at the origin and a Gaussian
distribution in the orthogonal direction centered at the
corresponding origin whose covariance matrix is scalar with $\sigma
= 5\%$ of the diameter of the cube, i.e., $2 \cdot \sqrt{d}$. Then,
for each subspace 250 samples are generated according to the
distribution just described. Next, the data is further corrupted
with 5\% or 30\% uniformly distributed outliers in a cube of
sidelength determined by the maximal distance of the former 250
samples to the origin (using the same code). The mean (along 100
instances) misclassification rate of the various algorithms is
recorded in Table~\ref{tab:error}, and the corresponding standard
deviation in Table \ref{tab:var}. The mean running time is shown in
Table~\ref{tab:time}.

From Table \ref{tab:error} we can see that MKF performs well in
various instances of hybrid linear modeling (with linear subspace),
and its advantage is especially obvious with many outliers
and high dimensions. 
The initialization of MKF with Algorithm \ref{alg:init} does not
work as well as random initialization. This is probably because both
the noise level and the outlier percentage are too large for the
former initialization, which is based on only a few nearest
neighbors. Nevertheless, we still notice that this initialization
reduces the running time of both KF and MKF.

We conclude from Table \ref{tab:time} that the running time of the
MKF algorithm is not as sensitive to the size of dimensions (either
ambient or intrinsic) as the running time of some other algorithms
such as GPCA, LSA and LSCC.

Table \ref{tab:var} indicates that GPCA and MoPPCA usually have a
larger standard deviation of misclassification rate, whereas other
algorithms have a smaller and comparable such standard deviation,
and are thus more stable. However, applying either KF or MKF without
restarts would result in large standard deviation of
misclassification rates due to convergence to local minima.

\subsection{Applications}

We apply the MKF algorithm to the Hopkins 155 database of motion
segmentation~\cite{Tron2007}, which is available at
http://www.vision.jhu.edu/data/hopkins155. This data contains 155
video sequences along with the coordinates of certain features
extracted and tracked for each sequence in all its frames.
The main task is to cluster the feature
vectors (across all frames) according to the different moving
objects and background in each video.

More formally, for a given video sequence, we denote the number of
frames by $F$. In each sequence, we have either one or two
independently moving objects, and the background can also move due
to the motion of the camera. We let $K$ be the number of moving objects
plus the background, so that $K$ is 2 or 3 (and distinguish
accordingly between two-motions and three-motions). For each
sequence, there are also $N$ feature points
$\by_1,\by_2,\cdots,\by_N \in \mathbb{R}^3$ that are detected on the
objects and the background. Let $\bz_{ij}\in \mathbb{R}^2$ be the
coordinates of the feature point $\by_j$ in the $i^{th}$ image frame
for every $1\leq i\leq F$ and $1\leq j \leq N$. Then
$\bz_j=[\bz_{1j},\bz_{2j},\cdots,\bz_{Fj}] \in \mathbb{R}^{2F}$ is
the trajectory of the $j^{th}$ feature point across the $F$ frames. The
actual task of motion segmentation is to separate these trajectory
vectors $\bz_1,\bz_2,\cdots,\bz_N$ into $K$ clusters representing
the $K$ underlying motions.

It has been shown~\cite{Costeira98} that under affine camera models
and with some mild conditions, the trajectory vectors corresponding
to different moving objects and the background across the $F$ image
frames live in distinct linear subspaces of dimension at most four
in $\mathbb{R}^{2F}$. Following this theory, we implement both the
MKF and KF algorithms with $d=4$.

\begin{table*}[htbp]
\centering \caption{\label{tab:comp1}  The mean and median
percentage of misclassified points for two-motions in Hopkins 155
database. We use 5 restarts for MKF and 20 for KF, and the smallest
of the $\ell_2$ errors is used. By MKF(R) and KF(R) we mean the
corresponding algorithm with random initialization.} \vspace{.1in}
\begin{tabular}{|l||r|r||r|r||r|r||r|r|}
  \hline

    &\multicolumn{2}{c||}{Checker}
    &\multicolumn{2}{c||}{Traffic}
    &\multicolumn{2}{c||}{Articulated}
    &\multicolumn{2}{c|}{All}\\
    \cline{2-9}
    \raisebox{1.5ex}[0pt]{\normalsize{2-motion}} &Mean &Median &Mean &Median &Mean &Median &Mean &Median\\
  \hline\hline
  CCS&16.37&10.64&5.27&0.00&17.58&7.07&12.16&0.00\\

  GPCA     & 6.09&1.03&1.41&0.00&2.88&0.00&4.59&0.38   \\
KF&5.33&0.04&2.36&0.00&3.83&1.11&4.43&0.00\\
KF 4$K$&5.81&0.17&3.55&0.02&4.97&1.15&5.15&0.06\\

KF 5&11.35&5.47&4.57&1.43&12.47&5.54&9.70&3.65\\
KF(R)&15.37&6.96&15.93&8.61&12.73&6.63&15.27&7.29\\

LLMC $4K$&4.65&0.11&3.65&0.33&5.23&1.30&4.44&0.24\\

LLMC 5&4.37&0.00&0.84&0.00&6.16&1.37&3.62&0.00\\
LSA $4K$&  2.57&0.27&5.43&1.48&4.10&1.22&3.45&0.59\\
LSA 5&8.84&3.43&2.15&1.00&4.66&1.28&6.73&1.99\\

\textbf{MKF}&3.70&0.00&0.90&0.00&6.80&0.00&3.26&0.00\\
MKF $4K$&4.51&0.01&1.59&0.00&6.08&0.92&3.90&0.00\\

MKF 5&9.37&4.10&3.47&0.00&10.68&5.84&7.97&2.39\\
MKF(R)&29.06&31.34&16.78&12.49&25.55&27.54&25.57&28.31\\
MSL&4.46&0.00&2.23&0.00&7.23&0.00&4.14&0.00\\

RANSAC&6.52&1.75&2.55&0.21&7.25&2.64&5.56&1.18\\

  \hline

\end{tabular}
\end{table*}

\begin{table*}[htbp]
\centering \caption{\label{tab:comp2}  The mean and median
percentage of misclassified points for three-motions in Hopkins 155
database. We use 5 restarts for MKF and 20 for KF, and the smallest
of the $\ell_2$ errors is used. And by MKF(R) and KF(R) we mean the
corresponding algorithm with random initialization.} \vspace{.1in}
\begin{tabular}{|l||r|r||r|r||r|r||r|r|}
  \hline

    &\multicolumn{2}{c||}{Checker}
    &\multicolumn{2}{c||}{Traffic}
    &\multicolumn{2}{c||}{Articulated}
    &\multicolumn{2}{c|}{All}\\
    \cline{2-9}
    \raisebox{1.5ex}[0pt]{\normalsize{3-motion}}  &Mean &Median &Mean &Median &Mean &Median &Mean &Median\\
  \hline\hline
  CCS& 28.63&33.21&3.02&0.18&44.89&44.89&26.18&31.74\\
GPCA&  31.95&32.93&19.83&19.55&16.85&28.66&28.66&28.26\\
KF&15.61&11.26&5.63&0.57&13.55&13.55&13.50&6.53\\

KF 4$K$&16.12&11.37&7.06&0.75&16.66&16.66&14.34&7.11\\

KF 5&26.95&31.88&8.09&5.67&17.65&17.65&22.65&25.08\\
KF(R)&21.83&24.52&8.70&5.00&15.85&15.85&18.86&17.81\\
LLMC $4K$&12.01&9.22&7.79&5.47&9.38&9.38&11.02&6.81\\

LLMC 5&10.70&9.21&2.91&0.00&5.60&5.60&8.85&3.19\\
LSA $4K$&5.80&1.77&25.07&23.79&7.25&7.25&9.73&2.33\\
LSA 5&30.37&31.98&27.02&34.01&23.11&23.11&29.28&31.63\\

\textbf{MKF}&14.50&12.00&3.06&0.01&15.90&15.90&12.29&6.23\\
MKF $4K$&14.26&10.85&3.17&0.00&15.68&15.68&12.12&5.02\\

MKF 5&24.77&25.85&9.47&5.82&21.19&21.19&21.51&21.39\\
MKF(R)&41.17&41.69&21.38&17.19&41.36&41.36&37.22&39.58\\

MSL&10.38&4.61&1.80&0.00&2.71&2.71&8.23&1.76\\

RANSAC&25.78&26.01&12.83&11.45&21.38&21.38&22.94&22.03\\

  \hline
\end{tabular}
\end{table*}

We compare the MKF with the following algorithms: Connected
Component Search (CCS)~\cite{LLMC}, improved GPCA for motion
segmentation (GPCA)~\cite{gpca_motion_08}, $K$-flats
(KF)~\cite{Ho03} (implemented for linear subspaces), Local Linear
Manifold Clustering (LLMC)~\cite{LLMC}, Local Subspace Analysis
(LSA)~\cite{Yan06LSA},
Multi Stage Learning (MSL)~\cite{Sugaya04}, and Random Sample
Consensus
(RANSAC)~\cite{Fischler81RANSAC,Torr98geometricmotion,Tron2007}.

We only directly applied KF and MKF, while for the other algorithms, we copy the results from
http://www.vision.jhu.edu/data/hopkins155 (they are based on
experiments reported in~\cite{Tron2007} and~\cite{LLMC}).

Since the database contains 155 data sets, we just record the mean
misclassification rate and the median misclassification rate for
each algorithm for any fixed $K$ (two or three-motions) and for the
different type of motions (``checker", ``traffic" and
``articulated") as well as the total database.

We use 5 restarts for MKF and 20 restarts for KF and record the best
segmentation result (both based on mean squared error). For MKF we
use the default value of $\di t=0.01$. Due to the randomness of both
MKF and KF, we applied them 100 times and recorded the mean and
 median of misclassification rates for both two-motions and
three-motions (see Table~\ref{tab:comp1} and Table~\ref{tab:comp2}).
We first applied both KF and MKF to the full data (with ambient
dimension $2F$). We applied KF and MKF with the initialization of
Algorithm~\ref{alg:init} as well as random initialization (and then
used the notation KF(R) and MKF(R)). For the purpose of comparison
with other algorithms (who could not be applied to the full
dimension), we also apply both KF and MKF to the data with reduced
dimensions: 5 and $4K$ (obtained by projecting onto the subspace
spanned by the top 5 or $4K$ right vectors of SVD). We denote the
corresponding application by KF 5, MKF 5, KF $4K$ and MKF $4K$. The
same naming convention was used for LSA and LLMC.
Table~\ref{tab:comp1} and Table~\ref{tab:comp2} report the results
for two-motions and three-motions respectively.

From Tables~\ref{tab:comp1} and~\ref{tab:comp2} we can see that MKF
(with the initialization of Algorithm~\ref{alg:init}) works well for
the given data. In particular, it exceeds the performance of many
other algorithms, despite that they are more complex. The clear
advantage of the initialization of Algorithm~\ref{alg:init} is
probably due to the cleanness of the data. It is interesting that
even though the data has low intrinsic dimensions, little noise and
few outliers, MKF is still superior to KF. This might be due to
better convergence of the MKF algorithm to a global minimum of the
$\ell_1$ energy, whereas KF might get trapped in a local and
non-global minimum more often.

The error rates of MKF and KF are very stable. Indeed, the standard
deviation of misclassification rate from MKF is always less than
0.002 for two-motions and less than 0.013 for three-motions.

\section{Conclusion and future work}
\label{sec:conclusions} We have introduced the Median $K$-flats
which is an online algorithm aiming to approximate a data set by $K$
best $\ell_1$ $d$-flats. It is implemented with a stochastic
gradient descent procedure which is experimentally fast. The
computational complexity is of order $O(n_s \cdot K \cdot d \cdot
D+n_s \cdot d^2 \cdot D)$ where $n_s$ is the number of sampling
iterations (typically about $10^4$, where for all experiments
performed here it did not exceed $3 \cdot 10^4$), and storage of the
MKF algorithm is of order $O(K \cdot d \cdot D)$. This algorithm
performs well on synthetic and real data distributed around mixtures
of linear subspaces of the same dimension $d$. It has a clear
advantage over other studied methods when the data has a large
component of outliers and when the intrinsic dimension $d$ is large.

There is much work to be done. First of all, there are many possible
practical improvements of the algorithm. In particular, we are
interested in extending the MKF algorithm to affine subspaces by
avoiding the normalization to the unit sphere (while incorporating
the necessary algebraic manipulations) as well as improving the
expected problematic convergence to the global minimum (due to many
local minima in the case of affine subspaces) by better
initializations. We are also interested in exploring methods for
determining the number of clusters, $K$, the intrinsic dimension,
$d$, and also developing strategies for mixed dimensions.

Second of all, we would like to pursue further applications of MKF.
For example, we believe that it can be used advantageously for
semi-supervised learning in the setting of hybrid linear modeling.
We would also like to exploit its ability to deal with both
substantially large and streaming data.

Third of all, it will also be interesting to try to comparatively
analyze the convergence of the following algorithms: MKF to the
global minimum of the $\ell_1$ energy of Eq.~\eqref{eq:l1energy}, a
straightforward $\ell_1$ version of the $K$-flats algorithm
(assuming an accurate algorithm for finding $\ell_1$ flats) to the
global minimum of the same energy, and $K$-flats to the global
minimum of the $\ell_2$ energy.

Last of all, we are currently developing a theoretical framework
justifying the robustness of $\ell_1$ minimization for many
instances of our setting. This theory also identifies some cases
where $\ell_1$ flats are not robust to outliers and careful
initializations are necessary for MKF.

\section*{Acknowledgment}
Thanks to the anonymous reviewers for their helpful comments,
Guangliang Chen for his careful reading and constructive comments on
earlier versions of this manuscript, and to Yi Ma and Rene Vidal for
guiding us through the literature on hybrid linear modeling and
responding to our many questions on the subject. We also thank Rene
Vidal for bringing to our attention the Hopkins 155 database as a
benchmark for motion segmentation algorithms. Thanks to the IMA, in
particular Doug Arnold and Fadil Santosa, for an effective
hot-topics workshop on multi-manifold modeling that we all
participated in. GL was supported by NSF grants \#0612608 and
\#0915064 (the latter one partially supported TZ), and AS was
supported by NSF grant \#0811203.

{\small
\bibliographystyle{ieee}
\bibliography{myrefs}

\begin{thebibliography}{10}\itemsep=-1pt

\bibitem{Basri03}
R.~Basri and D.~Jacobs.
\newblock Lambertian reflectance and linear subspaces.
\newblock {\em IEEE Transactions on Pattern Analysis and Machine Intelligence},
  25(2):218--233, February 2003.

\bibitem{Bishop:Pattern}
C.~M. Bishop.
\newblock {\em Pattern Recognition and Machine Learning (Information Science
  and Statistics)}.
\newblock Springer, August 2006.

\bibitem{Bradley00kplanes}
P.~Bradley and O.~Mangasarian.
\newblock k-plane clustering.
\newblock {\em J. Global optim.}, 16(1):23--32, 2000.

\bibitem{spectral_applied}
G.~Chen and G.~Lerman.
\newblock Spectral curvature clustering ({SCC}).
\newblock {\em Int. J. Comput. Vision}, 81(3):317--330, 2009.

\bibitem{Costeira98}
J.~Costeira and T.~Kanade.
\newblock A multibody factorization method for independently moving objects.
\newblock {\em International Journal of Computer Vision}, 29(3):159--179, 1998.

\bibitem{Fischler81RANSAC}
M.~Fischler and R.~Bolles.
\newblock Random sample consensus: A paradigm for model fitting with
  applications to image analysis and automated cartography.
\newblock {\em Comm. of the ACM}, 24(6):381--395, June 1981.

\bibitem{LLMC}
A.~Goh and R.~Vidal.
\newblock Segmenting motions of different types by unsupervised manifold
  clustering.
\newblock {\em Computer Vision and Pattern Recognition, IEEE Computer Society
  Conference on}, 0:1--6, 2007.

\bibitem{Ho03}
J.~Ho, M.~Yang, J.~Lim, K.~Lee, and D.~Kriegman.
\newblock Clustering appearances of objects under varying illumination
  conditions.
\newblock In {\em Proceedings of International Conference on Computer Vision
  and Pattern Recognition}, volume~1, pages 11--18, 2003.

\bibitem{Jain88}
A.~K. Jain and R.~C. Dubes.
\newblock {\em Algorithms for clustering data}.
\newblock Prentice-Hall, Inc., Upper Saddle River, NJ, USA, 1988.

\bibitem{Kambhatla94fastnon-linear}
A.~Kambhatla and T.~Leen.
\newblock Fast non-linear dimension reduction.
\newblock In {\em Advances in Neural Information Processing Systems 6}, pages
  152--159, 1994.

\bibitem{Kanatani01}
K.~Kanatani.
\newblock Motion segmentation by subspace separation and model selection.
\newblock In {\em Proc.~of 8th ICCV}, volume~3, pages 586--591. Vancouver,
  Canada, 2001.

\bibitem{Kanatani02}
K.~Kanatani.
\newblock Evaluation and selection of models for motion segmentation.
\newblock In {\em 7th ECCV}, volume~3, pages 335--349, May 2002.

\bibitem{Ma07Compression}
Y.~Ma, H.~Derksen, W.~Hong, and J.~Wright.
\newblock Segmentation of multivariate mixed data via lossy coding and
  compression.
\newblock {\em IEEE Transactions on Pattern Analysis and Machine Intelligence},
  29(9):1546--1562, September 2007.

\bibitem{Ma07}
Y.~Ma, A.~Y. Yang, H.~Derksen, and R.~Fossum.
\newblock Estimation of subspace arrangements with applications in modeling and
  segmenting mixed data.
\newblock {\em SIAM Review}, 50(3):413--458, 2008.

\bibitem{MacQueen67}
J.~MacQueen.
\newblock Some methods for classification and analysis of multivariate
  observations.
\newblock In {\em Proceedings of the 5th Berkeley Symposium on Mathematical
  Statistics and Probability}, volume~1, pages 281--297. University of
  California Press, Berkeley, CA, 1967.

\bibitem{Sugaya04}
Y.~Sugaya and K.~Kanatani.
\newblock Multi-stage unsupervised learning for multi-body motion segmentation.
\newblock {\em IEICE Transactions on Information and Systems},
  E87-D(7):1935--1942, 2004.

\bibitem{Tipping99mixtures}
M.~Tipping and C.~Bishop.
\newblock Mixtures of probabilistic principal component analysers.
\newblock {\em Neural Computation}, 11(2):443--482, 1999.

\bibitem{Torr98geometricmotion}
P.~H.~S. Torr.
\newblock Geometric motion segmentation and model selection.
\newblock {\em Phil. Trans. R. Soc. Lond. A}, 356:1321--1340, 1998.

\bibitem{Tron2007}
R.~Tron and R.~Vidal.
\newblock A benchmark for the comparison of 3-d motion segmentation algorithms.
\newblock In {\em CVPR}, 2007.

\bibitem{Tseng00nearest}
P.~Tseng.
\newblock Nearest $q$-flat to $m$ points.
\newblock {\em Journal of Optimization Theory and Applications},
  105(1):249--252, April 2000.

\bibitem{Vidal05}
R.~Vidal, Y.~Ma, and S.~Sastry.
\newblock Generalized principal component analysis ({GPCA}).
\newblock {\em IEEE Transactions on Pattern Analysis and Machine Intelligence},
  27(12), 2005.

\bibitem{gpca_motion_08}
R.~Vidal, R.~Tron, and R.~Hartley.
\newblock Multiframe motion segmentation with missing data using
  powerfactorization and gpca.
\newblock {\em Int. J. Comput. Vision}, 79(1):85--105, 2008.

\bibitem{Yan06LSA}
J.~Yan and M.~Pollefeys.
\newblock A general framework for motion segmentation: Independent,
  articulated, rigid, non-rigid, degenerate and nondegenerate.
\newblock In {\em ECCV}, volume~4, pages 94--106, 2006.

\bibitem{Yang_voting}
A.~Y. Yang and R.~M. Fossum.
\newblock Hilbert functions and applications to the estimation of subspace
  arrangements.
\newblock In {\em ICCV '05: Proceedings of the Tenth IEEE International
  Conference on Computer Vision (ICCV'05) Volume 1}, pages 158--165,
  Washington, DC, USA, 2005. IEEE Computer Society.

\end{thebibliography}
}

\end{document}